\documentclass{article}

    \usepackage[final]{neurips_2024}

\usepackage[utf8]{inputenc} 
\usepackage[T1]{fontenc}    
\usepackage{hyperref}       
\usepackage{url}            
\usepackage{booktabs}       
\usepackage{amsfonts}       
\usepackage{nicefrac}       
\usepackage{microtype}      
\usepackage{xcolor}         
\usepackage{afterpage}
\usepackage{natbib}
\setcitestyle{numbers,square}
\hypersetup{
	colorlinks=true,
	citecolor=teal,
}

\usepackage{amsmath,amsfonts,bm}

\def\eqref#1{equation~\ref{#1}}

\def\1{\bm{1}}

\DeclareMathAlphabet{\mathsfit}{\encodingdefault}{\sfdefault}{m}{sl}
\SetMathAlphabet{\mathsfit}{bold}{\encodingdefault}{\sfdefault}{bx}{n}

\definecolor{mycyan}{RGB}{212, 239, 251}
\usepackage{colortbl}
\usepackage{xcolor}
\definecolor{mygray}{gray}{.9}
\definecolor{goldenrod}{RGB}{245,245,220}
\newlength\savewidth\newcommand\shline{\noalign{\global\savewidth\arrayrulewidth\global\arrayrulewidth 1pt}\hline\noalign{\global\arrayrulewidth\savewidth}}
\newcolumntype{a}{>{\columncolor{mygray}}c}
\usepackage{fontawesome5}
\usepackage{booktabs}
\usepackage{makecell}
\usepackage{fontawesome5}
\usepackage{lipsum}
\usepackage{comment}
\usepackage{multirow}
\usepackage{wrapfig}
\usepackage{algorithm}
\usepackage{algorithmic}
\usepackage{xfrac}  
\usepackage{subfigure}

\usepackage{graphicx}
\usepackage{color}

\usepackage[english]{babel}
\usepackage{amsthm}
\definecolor{darkgreen}{rgb}{0,0.7,0}

\definecolor{mygraytext}{gray}{.75}

\title{Unveiling the Tapestry of Consistency in \\ Large Vision-Language Models}

\author{
  \hspace{-0.25cm}Yuan Zhang$^{1, 2}$, Fei Xiao$^{2}$, Tao Huang$^{3}$, Chun-Kai Fan$^{1}$, \textbf{Hongyuan Dong$^{2}$}, \\
  \hspace{-0.25cm}\textbf{Jiawen Li$^{2}$, Jiacong Wang$^{2, 4}$, Kuan Cheng$^{1}$, Shanghang Zhang$^{1}\thanks{Correspondence to: Shanghang Zhang and Haoyuan Guo.}$, \hspace{0.1cm}Haoyuan Guo$^{2 *}$} \\
 \hspace{-0.25cm}$^1$ School of Computer Science, Peking University \thanks{State Key Laboratory of Multimedia Information Processing.} \quad $^2$ ByteDance Inc\\
 \hspace{-0.25cm}$^3$ The University of Sydney \quad
 \hspace{-0.25cm}$^4$ School of Artificial Intelligence, UCAS\\
 \url{https://github.com/foundation-multimodal-models/ConBench}
}

\begin{document}

\maketitle

\begin{abstract}
Large vision-language models (LVLMs) have recently achieved rapid progress, exhibiting great perception and reasoning abilities concerning visual information.
However, when faced with prompts in different sizes of solution spaces, LVLMs fail to always give consistent answers regarding the same knowledge point. This inconsistency of answers between different solution spaces is prevalent in LVLMs and erodes trust.
To this end, we provide a multi-modal benchmark ConBench, to intuitively analyze how LVLMs perform when the solution space of a prompt revolves around a knowledge point.
Based on the ConBench tool, we are the first to reveal the tapestry and get the following findings:
(1) In the discriminate realm, the larger the solution space of the prompt, the lower the accuracy of the answers. 
(2) Establish the relationship between the discriminative and generative realms: the accuracy of the discriminative question type exhibits a strong positive correlation with its Consistency with the caption.
(3) Compared to open-source models, closed-source models exhibit a pronounced bias advantage in terms of Consistency.
Eventually, we ameliorate the consistency of LVLMs by trigger-based diagnostic refinement, indirectly improving the performance of their caption. We hope this paper will accelerate the research community in better evaluating their models and encourage future advancements in the consistency domain.
\end{abstract}  

\section{Introduction}

\begin{figure}[t]
    \vspace{-1mm}
    \centering
    \includegraphics[width=1\textwidth]{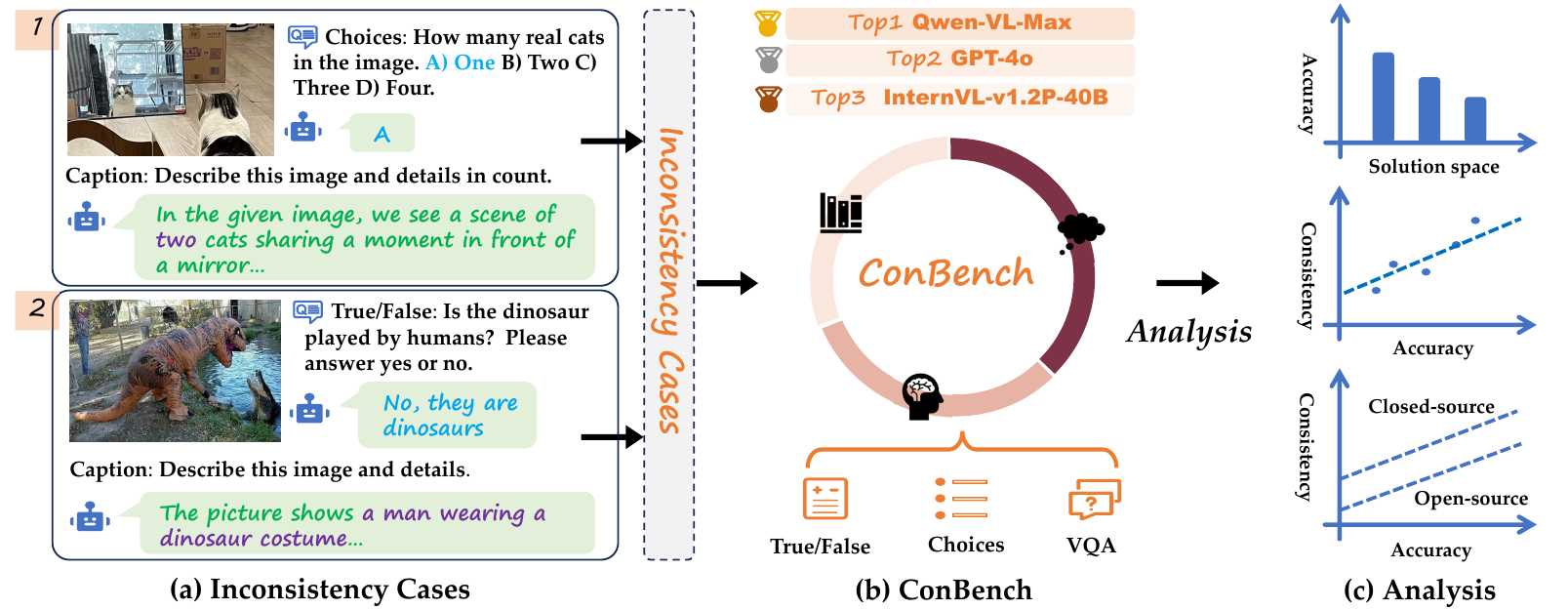}
    \vspace{-4mm}
    \caption{\textbf{Here is the overview of our paper.} Part (a) indicates two examples of Inconsistency between discriminative answers and generative captions, where the answers marked in blue contradict the answers marked in purple. Part (b) shows the Consistency evaluation method Conbench and its discriminative top three leaderboard. Part(c) reveals the main three findings built upon ConBench.}
    \label{fig:Moti}
    \vspace{-6mm}
\end{figure}


Recently, benefiting from notable advancements in large language models (LLMs)~\cite{achiam2023gpt, touvron2023llama, bai2023qwen}, the realm of large vision-language models (LVLMs) has undergone a revolutionary transformation. These novel LVLMs \cite{liu2023improvedllava,team2023gemini, Qwen-VL, chen2023internvl, li2024mini, li2023blip} try to combine visual signals with textual semantics and spark cognitive brilliance across modalities. 
Although LVLMs can generate high-quality responses to task prompts, we discover that for correctly answered cases, simply modifying the prompt will result LVLMs in providing contradictory responses. In Figure \ref{fig:Moti} (a.2), LLaVA-7B \cite{liu2023improvedllava} properly describes the picture as “\verb+It is a man wearing a dinosaur costume.+”, but when prompted “\verb+Is the dinosaur played by humans? Please answer yes or no.+”, it responds with “\verb+No, they are dinosaurs+”. The above phenomenon of Inconsistency is widely observed across mainstream LVLMs, and a preliminary study was conducted only on LLMs \cite{li2023benchmarking}. In practice, in contrast to the fixed patterns of questions, designed in existing multimodal benchmarks, the users tend to pose questions in arbitrary ways. Therefore, it is necessary to ensure the LVLMs in predicting correct and \textbf{consistent} answers, even when faced with various formats of queries.

However, there are currently no benchmarks or research studies that specifically focus on evaluating the Consistency of LVLMs responses. These single-prompt type evaluation approaches \cite{li2023seed,fu2023mme,yue2023mmmu,liu2023mmbench,chen2024we} lead to a disconnect between benchmark accuracy and real-world user practical experience.

Based on the above observations, we systematically introduce a \textbf{Con}sistency \textbf{Bench}mark dubbed ConBench, to estimate the capabilities of LVLMs more thoroughly via diverse question formats. It consists of $1,000$ public pictures, and each was manually selected from four multimodal benchmarks \cite{fu2023mme, li2023seed, yue2023mmmu, liu2023mmbench}. Apart from the original discriminative prompt, we constructed two additional discriminative types of questions\footnote{e.g., Multiple-choice questions and limited VQA questions are generated for MME benchmark.} by ChatGPT/GPT-4 \cite{achiam2023gpt}. Notably, three types of questions of each case are around the \textbf{same} knowledge point. Besides, every set is accompanied by a generative question without ground truth. Consequently, ConBench serves as an evaluation tool that observes the Consistency performance of LVLMs and surpasses the limitations of previous assessments.

Furthermore, grounded on the ConBench, we conduct an in-depth analysis and visualization of Consistency on $14$ popular LVLMs. In a nutshell, the conclusions of noteworthy insight are threefold:

\textbf{C1} In the \textbf{discriminative} question-answering (QA) domain: (1) A decrease in LVLMs accuracy as the prompt's solution space increases. (2) Instances of erroneous yet consistent answers are scarce.

\textbf{C2} Extended to the \textbf{generative} domain, we establish a connection between discriminative and generative domains by the perspective of Consistency. (1) As the solution space of discriminative questions expands, the Consistency between its answer and caption grows stronger. (2) The accuracy of discriminative answer and its Consistency with the caption exhibit a positive correlation.

\textbf{C3} Closed-source models exhibit a pronounced bias advantage in terms of Consistency, compared to open-source models. This provides an alternative perspective to demonstrate why closed-source models, despite sometimes having lower accuracy, offer a better user experience in practical applications.

Eventually, leveraging the insights gained from our theoretical discoveries, we enhance the caption performance of LVLMs without any additional costs associated with training. Specifically, we construct discriminative prompts based on the low-confidence words in the answers of LVLMs, forcing the LVLMs to introspect. Then, through iterative refinement in multiple rounds of question-answering, the quality of LVLMs' captions gets an impressive achievement (e.g., our method improves the LLaVA-NeXT-34B \cite{liu2024llavanext} by $9.1\%$ and MiniGemini-34B \cite{li2024mini} by $9.6\%$ on metric[C] in Sec. \ref{metric}).

In summary, to the best of our knowledge, we are the first to propose a Consistency evaluation method and conduct a comprehensive analysis of Inconsistency in LVLMs. We hope this paper serves as a catalyst for further exploration, and look forward to the community applying the above findings to polish up the usability and practicality of large vision-language models.  

\section{Related Work}

\paragraph{Large Vison Language Models}
With the impressive success of large language models (LLMs) \cite{achiam2023gpt, touvron2023llama, bai2023qwen, bi2024deepseek, zhang2023llama}, recent studies work on generative large vision-language models (LVLMs) \cite{liu2023improvedllava, team2023gemini, Qwen-VL, chen2023internvl, li2024mini, ye2023mplug} to improve multimodal comprehension and generation through utilizing the strong generality of LLMs. Built upon the CLIP \cite{radford2021learning} image encoder which is somewhat aligned with the language modality, current LVLMs typically utilize vast image-text pairs to connect the vision encoder and LLM, enabling LLM to receive and understand visual content. 
For instance, LLaVA \cite{liu2023llava} directly connects the vision encoder and LLM with MLPs, showing proficiency in multi-modal dialogues. Subsequent works have further enhanced LVLMs by improving the multi-modal instruction data \cite{liu2023improvedllava, ye2023mplug, chen2023sharegpt4v} and designing novel modules \cite{Qwen-VL, bi2024deepseek, wang2023cogvlm} for more sufficient modality alignment.

\paragraph{Conventional Multimodal Evaluation}
A multitude of public multimodal benchmarks, such as MME \cite{fu2023mme}, SeedBench \cite{li2023seed}, and MMBench \cite{liu2023mmbench}, further advance objective evaluation of LVLMs by only constructing True/False questions or multiple-choice questions, where the absence of diverse question types causes instability. In addition, their objective metrics solely emphasize the LVLM's accuracy, disregarding its robustness and security. The above issues can lead to a situation where some LVLMs have lower accuracy in evaluation results but provide a better user experience. To systematically assess the comprehensive capability of LVLMs, we propose a simple and efficient evaluation approach that relies on checking the Consistency between different kinds of prompts.

\paragraph{Inconsistency in LLMs}
A amount of prior work has been conducted on investigating Inconsistency in LLMs. \cite{li2023benchmarking} is the first to find the Inconsistency phenomenon in question-answering and validator tasks and define GV-consistency. Besides, it leverages consistency pair for training to improve LLMs' performance. While \cite{lin2023generating} utilizes Consistency to check for hallucination detection in LLMs, a logic consistency-based
method that involves logic-related questions and answers. Compared to LLMs, Inconsistency in LVLMs is more likely to occur due to the additional visual modality, which deserves further exploration.

\definecolor{mycyan}{RGB}{34,139,34}
\definecolor{mycyan2}{RGB}{0,120,200}
\definecolor{mycyan3}{RGB}{220,50,50}

\begin{wrapfigure}{r}{0.45\textwidth}
    \vspace{-18mm}
    \begin{center}
        \includegraphics[width=\linewidth]{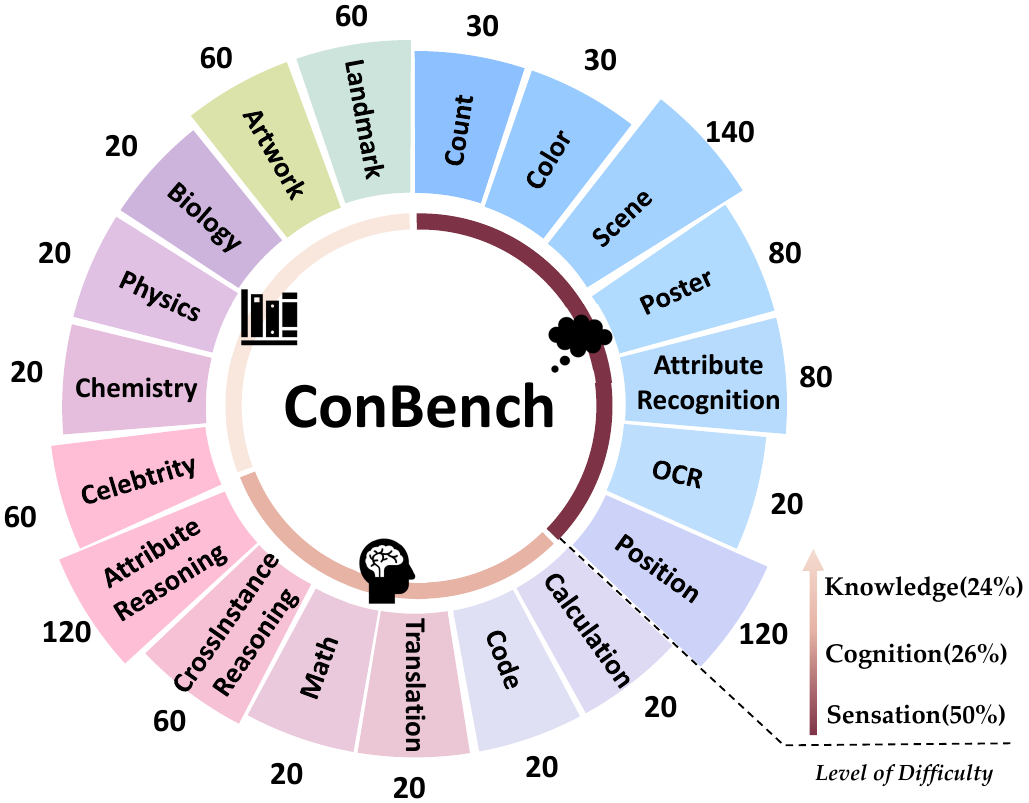}
    \end{center}
    \vspace{-4mm}
    \caption{\textbf{Overview of 19 evaluation detailed categories in ConBench.}}
    \label{fig:ConBench}
    \vspace{-10mm}
\end{wrapfigure}
\section{ConBench}
We propose a novel multimodal evaluation pipeline named ConBench to comprehensively assess LVLMs. The ConBench has a total of 4K questions on 1K images and corresponding 3K discriminative ground truths, guaranteeing evaluation quality in terms of the quantity and diversity of questions. In Sec. \ref{3.1}, we present the generation of ConBench and the construction pipeline for prompts. Sec. \ref{3.2} introduces the hierarchical core capabilities and discusses the design philosophy. Sec. \ref{3.3} and \ref{3.4} describe the evaluation strategy for scoring various types of answers.

\subsection{Data Generation Process}
\label{3.1}
\begin{wrapfigure}{r}{0.47\textwidth}
        \vspace{-7mm}
        \begin{center}
            \includegraphics[width=\linewidth]{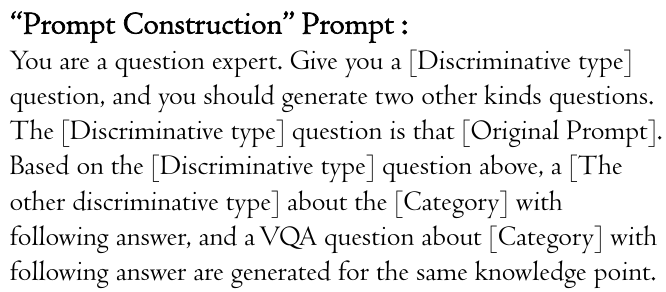}
        \end{center}
        \vspace{-4mm}
        \caption{\textbf{The prompt for generation of discriminative questions.} Please zoom in to view.}
        \label{fig:prompt}
        \vspace{-3mm}
    \end{wrapfigure}

    \paragraph{Image Filter}
    We manually chose 1K images from four high-quality multimodal benchmarks: MME \cite{fu2023mme}, SeedBench \cite{li2023seed}, MMBench \cite{liu2023mmbench}, and MMMU \cite{yue2023mmmu}. MME is a true/false question type, while SeedBench and MMBench cover comprehensive multiple-choice questions. Meanwhile, MMMU emphasizes the knowledge level. The criteria for the image filter include: (1) resolution is more than $224 \times 224$ (2) the image rarely occurs in the mainstream training dataset (e.g., COCO \cite{lin2014microsoft} and Cityscapes \cite{cordts2016cityscapes}) (3) There are more than 3 foreground objects in the image. The above criteria ensure the quality of content in images. 
    \vspace{-4mm}
    \paragraph{Prompt Construction}
    Each image is accompanied by its original discriminative prompt, and we constructed two extra discriminative questions. Therefore, a case owns three discriminative prompts (true/false, multiple-choice and limited VQA questions) with a generative caption prompt around the same knowledge point. Firstly, we modified the original prompts whose answers can be directly inferred from the text instead of the image, to force LVLMs to utilize information from the visual features. Next, we employed GPT/GPT-4 to generate the extra discriminative types of questions, which were then subjected to the manual review, and the proposed prompt is listed in Figure \ref{fig:prompt}.
    Finally, to avoid bias in the LVLMs that may affect the evaluation results, the true/false questions have a $50\%$ distribution for both correct and wrong ground truths. For the multiple-choice questions, each option (e.g., A, B, C, D) has an equal probability distribution of $25\%$ for being the correct answer. Notably, to ensure an accurate evaluation parser, limited VQA questions are subject to certain restrictions, like specifying the word count and answer format (e.g., fractions / abbreviations / numbers). 

\subsection{Hierarchical Core Capabilities}
\label{3.2}
    The ConBench comprises three core capabilities, arranged in ascending order of difficulty, namely: Sensation, Cognition, and Knowledge, with nineteen fine-grained dimensions shown in Figure \ref{fig:ConBench}. 
    \textcolor{mycyan}{\textbf{[Easy Mode]}} \textbf{Sensation: What you see is what you get.} We assume that sensation is the most fundamental expertise of LVLMs, and it is the "eye" of the LLMs. While perceived questions appear simple and basic, they are nonetheless essential. Therefore, this capability accounts for $50\%$ of the ConBench. Count, color, optical character recognition (OCR) and scene categories focus on subtle details, while poster, attribute recognition and position types emphasize the overall picture. 

    \textcolor{mycyan2}{\textbf{[Medium Mode]}} \textbf{Cognition: Go beyond the surface.} The cognitive process needs the model to integrate visual and language modalities: observing the content of an image, combining it with the text of question, and retrieving knowledge from within the LLMs. It is more challenging than the single sensation task. This section constitutes $26\%$ of the ConBench, including numerical calculation, code inference, text translation, math, cross-instance reasoning and attribute reasoning categories.
    
    \textcolor{mycyan3}{\textbf{[Hard Mode]}} \textbf{Knowledge: Master the art of synthesis and integration.} Mastering professional knowledge is an essential pathway for next-generation LVLMs to become Expert AGI, as it requires a higher level of understanding of images and the application of expert knowledge. We carefully selected knowledge from diverse and extensive fields, such as celebrities, chemistry, physics, biology, art and geography. This part takes up $24\%$ of the total, and functions as the upper limit of ConBench.

\subsection{Results Parser}
\label{3.3}
For true/false questions, we first extract the "yes" and "no" from the answer. If both of them are absent, the answer would be considered as "none". Then, we strictly compare the extracted answer with the ground truth. If they match exactly, the true/false response is considered correct. 

When parsing the outcome of multiple choices, we derive the choice label (e.g., A, B, C, D) from it. If successful, utilize this as the prediction and match the ground truth. If not, we will not proceed with further extracting the answers. Because in each prompt of choices, we specified that only one letter needs to be answered. Doing so would be unfair to LVLMs that excel in following instructions.

We still utilize character matching for the answer of limited VQA instead of GPTs. On one hand, we have taken strict formatting constraints on the prompts. For instance, in physics and math, there are restrictions on answering with fractions (e.g., 1/2), while in geography at the city level. On the other hand, the cost of the GPT's judgment is high and the waiting time is delayed. Specifically, the parser is based on the Average Normalized Levenshtein Similarity (ANLS) \cite{mathew2021docvqa}, where the threshold $\tau$ is set to $0.95$ and $M = N = 1$. When parsed result $s > 0.4$, we consider the answer to be exactly right.

\subsection{Multidimensional Evaluation Metric}
\label{3.4}
Here we provide two evaluation metrics, each from the perspective of discriminative and generative domains, aiming to provide a more comprehensive understanding of LVLMs consistency. The former does not rely on AI tools and quickly produces Consistency results among discriminative responses via Sec. \ref{3.3}, primarily evaluating the knowledge. The latter employs GPT to indirectly assess the quality of captions, by judging the consistency between discriminative responses and captions.

\textbf{Discriminative Domain Evaluation Metric}
We define the \textbf{ConScore[D]} as that: when \textbf{all three discriminative} types of questions within the same case are answered correctly, the model gets one point. The maximum score is 1000 points. The final format is presented as a percentage (\%).

\begin{figure}[t]
    \centering
    \includegraphics[width=1\textwidth]{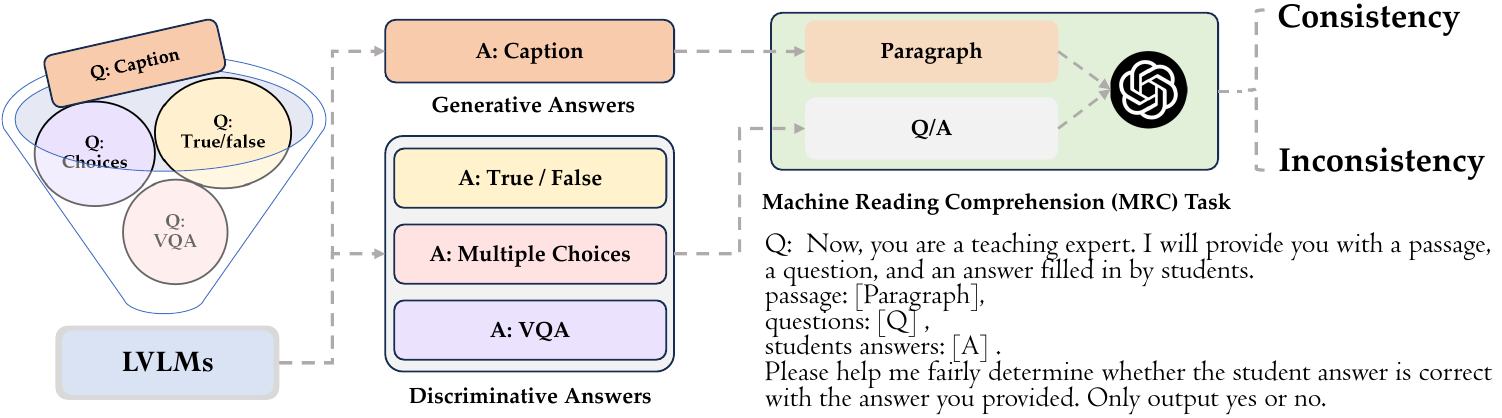}
    \vspace{-4mm}
    \caption{\textbf{The pipeline of judging Consistency between caption and discriminative answers via GPT/GPT4.} Please zoom in to view the prompt.}
    \label{fig:pipeline}
    \vspace{-2mm}
\end{figure}
\textbf{Generative Domain Evaluation Metric}
\label{metric}
Due to the high variability in captions, it is not possible to calculate Consistency based on character matching alone. Therefore, we rely on GPT/GPT4 for judgment. The judging process and the constructed prompts are shown in Figure \ref{fig:pipeline}. We formulate it as a machine reading comprehension task. We manually sample the judgment results, and GPT4 achieved an accuracy rate of $95\%$, which is reliable and trustworthy. Next, we define the \textbf{ConScore[C]} as the average score of Consistency between the caption and the other three discriminative responses.

\newtheorem{theorem}{Fact}[subsection]

\section{Analysis}
\subsection{Evaluation Results}

\begin{table}[t]
    \centering
	\renewcommand\arraystretch{1.5}
	\setlength\tabcolsep{0.55mm}
	\caption{\textbf{Evaluation[D] of mainstreams series of LVLMs on ConBench.} The detailed results of the Sensation, Cognition, and Knowledge core capabilities are listed below. T, C, and V represent true-false, multiple-choice, and limited VQA questions, respectively. The ranking can be found below the respective numbers. $\dagger$: \scriptsize{Due to safety considerations, GPT-4V declined to answer the celebrity category.}}
	\vspace{-2mm}
	\label{tab:leaderborad}
	\scriptsize
        \begin{center}
	\begin{tabular}{l|c|p{0.7cm}p{0.7cm}p{0.7cm}p{0.55cm}|p{0.7cm}p{0.7cm}p{0.7cm}p{0.55cm}|p{0.7cm}p{0.7cm}p{0.7cm}p{0.55cm}}
	    \shline
	    \multirow{2}{*}{\footnotesize{Method}} & \multirow{2}{*}{\footnotesize{ConScore[D]}} & \multicolumn{4}{c|}{\footnotesize{Sensation}} & \multicolumn{4}{c|}{\footnotesize{Cognition}} & \multicolumn{4}{c}{\footnotesize{Knowledge}} \\
          & & \multicolumn{1}{c}{\footnotesize{T}} & \multicolumn{1}{c}{\footnotesize{C}} & \multicolumn{1}{c}{\footnotesize{V}} & Con & \multicolumn{1}{c}{\footnotesize{T}} & \multicolumn{1}{c}{\footnotesize{C}} & \multicolumn{1}{c}{\footnotesize{V}} & Con & \multicolumn{1}{c}{\footnotesize{T}} & \multicolumn{1}{c}{\footnotesize{C}} & \multicolumn{1}{c}{\footnotesize{V}} & Con \\
	    \shline
            \rowcolor{mygray}
	    \multicolumn{14}{c}{\textit{Closed-source Vision Language Models}}\\
            GPT-4V$^\dagger$ \cite{achiam2023gpt} & $29.20_{6}$ & $80.4$ & $79.0$ & $61.7$ & $48.3$ & $68.8$ & $53.2$ & $39.9$ & $20.4$ & $63.1$ & $57.2$ & $30.0$ & $14.2$ \\
            GPT-4-Omni \cite{achiam2023gpt} & $\underline{35.70}_{2}$ & $89.2$ & $79.4$ & $64.4$ & $\underline{55.0}$ & $71.8$ & $62.8$ & $44.9$ & $27.8$ & $64.7$ & $61.7$ & $39.7$ & $\underline{23.3}$ \\
    	Gemini-Pro-Vision \cite{team2023gemini} & $25.00_{10}$ & $85.2$ & $60.7$ & $63.4$ & $39.3$ & $71.8$ & $45.0$ & $44.2$ & $15.1$ & $65.0$ & $51.4$ & $39.7$ & $15.8$ \\
            Gemini-Ultra-Vision \cite{team2023gemini} & $33.10_{4}$ & $78.9$ & $78.6$ & $66.3$ & $50.3$ & $68.1$ & $58.5$ & $47.9$ & $\underline{28.5}$ & $62.9$ & $62.2$ & $44.7$ & $19.7$ \\
            Qwen-VL-Plus \cite{Qwen-VL} & $28.10_{7}$ & $82.7$ & $74.9$ & $60.4$ & $45.0$ & $64.2$ & $41.7$ & $30.8$ & $16.3$ & $63.6$ & $54.2$ & $33.3$ & $15.8$ \\
            Qwen-VL-Max \cite{Qwen-VL} & $\mathbf{37.00}_{1}$ & $86.4$ & $80.7$ & $65.4$ & $\mathbf{56.3}$ & $72.9$ & $51.4$ & $51.3$ & $28.1$ & $68.3$ & $58.6$ & $38.9$ & $\mathbf{24.2}$ \\
	    \hline
     
            \rowcolor{mygray}
	    \multicolumn{14}{c}{\textit{7B Vision Language Models}}\\
            LLaVA-v1.5-7B \cite{liu2023improvedllava} & $16.60_{14}$ & $79.3$ & $56.8$ & $44.3$ & $28.3$ & $51.4$ & $33.5$ & $15.8$ & $4.7$ & $61.7$ & $44.4$ & $16.9$ & $7.8$ \\
            Qwen-VL-Chat \cite{Qwen-VL} & $26.40_{9}$ & $81.0$ & $79.6$ & $54.2$ & $39.0$ & $55.0$ & $46.3$ & $33.2$ & $13.5$ & $60.3$ & $54.2$ & $28.9$ & $14.7$ \\
	    \hline

            \rowcolor{mygray}
	    \multicolumn{14}{c}{$\sim$ \textit{13B Vision Language Models}}\\
            LLaVA-v1.5-13B \cite{liu2023improvedllava} & $24.00_{11}$ & $82.9$ & $77.1$ & $49.6$ & $39.5$ & $53.6$ & $37.8$ & $20.1$ & $10.4$ & $65.6$ & $50.3$ & $17.2$ & $9.7$ \\
            MiniGemini-13B \cite{li2024mini}  & $21.80_{13}$ & $81.9$ & $69.7$ & $52.8$ & $39.3$ & $51.9$ & $38.2$ & $21.1$ & $6.9$ & $52.8$ & $36.7$ & $17.5$ & $9.2$ \\
            InternVL-v1.5-26B \cite{chen2024far} & $31.40_{5}$ & $85.6$ & $84.8$ & $65.0$ & $54.3$ & $59.7$ & $58.6$ & $44.4$ & $19.4$ & $58.1$ & $55.8$ & $25.3$ & $12.2$ \\
	    \hline

            \rowcolor{mygray}
	    \multicolumn{14}{c}{$\sim$ \textit{34B Vision Language Models}}\\
            LLaVA-NeXT-34B \cite{liu2024llavanext} & $27.70_{8}$ & $82.4$ & $81.7$ & $55.6$ & $43.6$ & $50.7$ & $47.5$ & $25.6$ & $9.9$ & $60.4$ & $56.1$ & $27.8$ & $12.8$ \\
            MiniGemini-34B \cite{li2024mini} & $23.00_{12}$ & $80.8$ & $76.8$ & $48.2$ & $39.7$ & $36.9$ & $30.7$ & $18.9$ & $6.0$ & $58.1$ & $42.3$ & $20.8$ & $8.2$ \\
            InternVL-v1.2P-40B \cite{chen2023internvl} & $34.70_{3}$ & $83.7$ & $83.2$ & $66.6$ & $53.4$ & $74.2$ & $67.6$ & $57.1$ & $\mathbf{34.9}$ & $72.2$ & $58.3$ & $28.6$ & $13.6$ \\

	    \shline
	\end{tabular}
         \end{center}
    \hspace{2mm}
    \vspace{-7mm}
\end{table}
\definecolor{ForestGreen}{RGB}{34,139,34}
\newcommand{\fg}[1]{\mathbf{\mathcolor{ForestGreen}{#1}}}
\definecolor{Forestred}{RGB}{220,50,50}
\newcommand{\fr}[1]{\mathbf{\mathcolor{Forestred}{#1}}}
\begin{table}[t]
    \centering
	\renewcommand\arraystretch{1.45}
	\setlength\tabcolsep{0.55mm}
	\caption{\textbf{Evaluation of Consistency between caption and three discriminative types of answer on ConBench.} The "rank diff" means the difference between ConScore[D] and Score[C]. The Con[$X$] is the Consistency ratio between discriminative answer type $X$ and caption. The "ordered" represents whether Con[T] < Con[C] < Con[V] is in its line.}
	\vspace{-3mm}
	\label{tab:caption_board}
	\scriptsize
    \begin{center}
	\begin{tabular}{l|c|c|p{1.6cm}p{1.6cm}p{1.6cm}c}
	    \shline
        \footnotesize{Method} & \footnotesize{Rank Diff} & \footnotesize{ConScore[C]} & \footnotesize{Con[T]} & \footnotesize{Con[C]} & \footnotesize{Con[V]} & \footnotesize{Ordered}\\
	    \shline
            \rowcolor{mygray}
	    \multicolumn{7}{c}{\textit{Closed-source Vision Language Models}}\\
            GPT-4V \cite{achiam2023gpt} & $\fg{\uparrow3}$ & $55.6_{3}$ & $51.20$ & $56.50$ & $59.10$ & Y  \\
            GPT-4-Omni \cite{achiam2023gpt} & $\fg{\uparrow1}$ & $\mathbf{62.2}_{1}$ & $58.00$ & $62.50$ & $66.10$ & Y \\
            Gemini-Pro-Vision \cite{team2023gemini} & $\fg{\uparrow1}$ & $48.4_{9}$ & $43.30$ & $45.20$ & $56.80$ & Y \\
            Gemini-Ultra-Vision \cite{team2023gemini} & $\mathbf{-}$ & $54.6_{4}$ & $47.80$ & $55.20$ & $60.70$ & Y \\
            Qwen-VL-Plus \cite{Qwen-VL} & $\mathbf{-}$ & $50.2_{7}$ & $47.10$ & $49.10$ & $54.30$ & Y \\
            Qwen-VL-Max \cite{Qwen-VL}  & $\fr{\downarrow1}$ & $\underline {58.4}_{2}$ & $54.30$ & $58.00$ & $62.90$ & Y \\
	    \hline
     
            \rowcolor{mygray}
	    \multicolumn{7}{c}{\textit{7B Vision Language Models}}\\
            LLaVA-v1.5-7B \cite{liu2023improvedllava}  & $\mathbf{-}$ & $38.4_{14}$ & $39.20$ & $36.60$ & $39.50$ & N\\
            Qwen-VL-Chat \cite{Qwen-VL} & $\fr{\downarrow2}$  & $48.0_{11}$ & $42.00$ & $50.80$ & $51.30$ & Y\\
	    \hline

            \rowcolor{mygray}
	    \multicolumn{7}{c}{$\sim$ \textit{13B Vision Language Models}}\\
            LLaVA-v1.5-13B \cite{liu2023improvedllava} & $\fr{\downarrow1}$ & $44.4_{12}$ & $41.50$ & $45.80$& $46.00$ & Y\\
            MiniGemini-13B \cite{li2024mini}  & $\mathbf{-}$ & $41.7_{13}$ & $38.80$ & $42.90$ & $43.30$ & Y\\
            InternVL-v1.5-26B \cite{chen2024far} & $\fr{\downarrow1}$ & $50.9_{6}$ & $44.50$ & $53.90$& $54.20$ & Y\\
	    \hline

            \rowcolor{mygray}
	    \multicolumn{7}{c}{$\sim$ \textit{34B Vision Language Models}}\\
            LLaVA-NeXT-34B \cite{liu2024llavanext} & $\fr{\downarrow2}$ & $48.3_{10}$ & $46.00$ & $52.20$ & $46.80$ & N\\
            MiniGemini-34B \cite{li2024mini} & $\fg{\uparrow4}$ & $49.6_{8}$ & $56.80$ & $48.00$ & $44.10$ & N\\
            InternVL-v1.2P-40B \cite{chen2023internvl} & $\fr{\downarrow2}$ & $53.7_{5}$ & $49.80$ & $55.50$ & $55.80$ & Y\\

	    \shline
	\end{tabular}
         \end{center}
    \hspace{2mm}
    \vspace{-5mm}
\end{table}

In this section, 6 closed-source and 8 open-source representative LVLMs with varying sizes and architectures are evaluated on our Consistency benchmark, including GPT-4V \cite{achiam2023gpt}, GPT4-Omni \cite{achiam2023gpt}, Gemini-Vision \cite{team2023gemini}, Qwen-VL series \cite{Qwen-VL}, LLaVA series \cite{liu2023improvedllava, liu2024llavanext}, MiniGemini series \cite{li2024mini} and InternVL series \cite{chen2023internvl}. The evaluation results on ConBench are listed in Table \ref{tab:leaderborad} and \ref{tab:caption_board}.
In the metric[D] \ref{tab:leaderborad}, Qwen-VL-Max \cite{Qwen-VL} secures the top position, leading the second-place GPT4-Omni \cite{achiam2023gpt} by a margin of $1.3\%$. The InternVL-v1.2P-40B \cite{chen2023internvl} performs best in the open-sourced community, especially in cognition capability. The LLaVA series did not make it to the top ten.
In the metric[C], the newest GPT4-Omni \cite{achiam2023gpt} leads the leaderboard, which is the only model that surpasses $60$. It has a significant advantage over the second-place model Qwen-VL-Max \cite{Qwen-VL}, with a gap of $3.8$. We observed that although the GPT series slightly underperforms Qwen-Max in metric[D], it significantly outperforms the Qwen series in metric[C], which aligns with our actual user experience. Actually, ConScore[C] provides an alternative quality description of captions, because higher recall and precision rates usually match better Consistency. Besides, rankings of LVLMs show a slight variation between metric[C] and metric[D]. The GPT series models claim better performance of caption generation.

    \begin{wrapfigure}{r}{0.55\textwidth}
        \vspace{-7mm}
        \begin{center}
            \includegraphics[width=\linewidth]{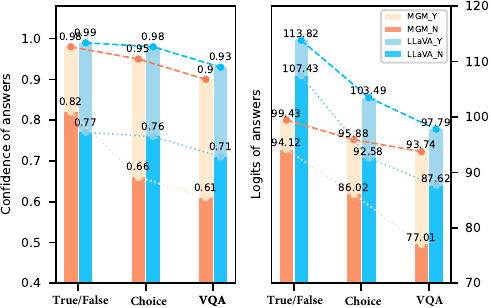}
        \end{center}
        \vspace{-4mm}
        \caption{The confidence and logits of answers of LLaVA-13B and MGM-13B.}
        \label{fig:conf}
        \vspace{-7mm}
    \end{wrapfigure}

\subsection{Discriminative Domain}
    To investigate what causes the Inconsistency between different types of prompts, we first conduct analyses on the discriminative domain to compare the performance differences. We summarize our findings into the following facts:

    \begin{theorem}[Inconsistency in Accuracy]
        \label{f1}
        The accuracy of the answer decreases as the solution space of the discriminative prompt increases.
    \end{theorem}
        As shown in the columns of "T", "C", and "V" in Table \ref{tab:leaderborad}, the accuracy decreases as the solution space expands in all core capabilities. For instance (e.g., the Sensation of GPT-4-Omni), the double-choice true-false questions achieve an accuracy of $89.2$, whereas the accuracy for multiple-choice and VQA questions on the same case declines to $79.4$ and $61.7$, respectively. This is understandable, as the number of potential choices increases, the difficulty in identifying the correct answer also rises.
        
    

    \begin{table}[t]
    \renewcommand\arraystretch{2.3}
    \setlength\tabcolsep{1.5mm}
    \centering
    \caption{\textbf{The Consistency between multiple choices and VQAs, including both correct and wrong.} Each case is picked up in order from top to bottom from the 14 LVLMs in Table \ref{tab:leaderborad}.}
    \vspace{-1mm}
    \label{tab:ab_2}
	\scriptsize
	\begin{tabular}{c|cccccccccccc}
	    \shline
            Con[Correct] (\%)& 35.00 & 39.90 & 31.20 & 34.10 & 41.60 & 37.60 & 39.40 & 29.30 & 39.70 & 25.90 & 28.70 & 22.20 \\
	    \shline
            Con[Wrong] (\%)& 0.30 & 0.40 & 0.20 & 0.50 & 0.40 & 0.50 & 0.30 & 0.40 & 0.20 & 0.50 & 0.10 & 0.20\\
            
	    \shline
	\end{tabular}
\end{table}    
    \begin{theorem}[Inconsistency in Wrong Answers]
        \label{f2}
        Cases of erroneous yet consistent answers are scarce.
    \end{theorem}
        We analyze the answers that fail in all three question types and find that, despite all resulting in incorrect predictions, they do not demonstrate a consistent understanding of the same images, leading to distinct answers. For example, we calculated the proportion of consistent incorrect responses in VQA and multiple-choice questions. We found a very small consistency, and it did not exceed $0.50\%$ across the entire benchmark. This indicates that the models struggle to interpret the visual content uniformly, revealing significant variability in their failure modes.
    
    \begin{theorem}[Inconsistency in Confidence]
        The confidence of models in their answers reveals signs of inconsistent and incorrect predictions.
    \end{theorem}
        
        Taking Fact \ref{f1} and Fact \ref{f2} into account, we perform a deeper analysis of the model's predictions by measuring their confidence in the answers. We use the predicted probabilities and logits of the answer tokens to represent confidence (see Appendix \ref{stat} for details). As summarized in Figure \ref{fig:conf}, we measure the average probabilities and logits of the correct and incorrect answers\footnote{MGM\_Y and LLaVA\_Y mean the correct, while MGM\_N and LLaVA\_N represent the wrong.}, respectively. The three types of questions share similar confidence levels for the correct answers. However, for the incorrect answers, their confidence levels vary significantly with a clear trend: the larger the solution space, the smaller the confidence. This analysis provides crucial insights for our method in enhancing the consistency and accuracy of LVLMs, which we will further discuss in Sec. \ref{improve}.

\subsection{Generative Domain}
    Next, we extend our attention to the generative domain. Based on Consistency, we first build a bridge between the discriminative and generative domains. We consolidate our findings as the below facts:

    \begin{figure}[t]
        \centering
        \includegraphics[width=1\textwidth]{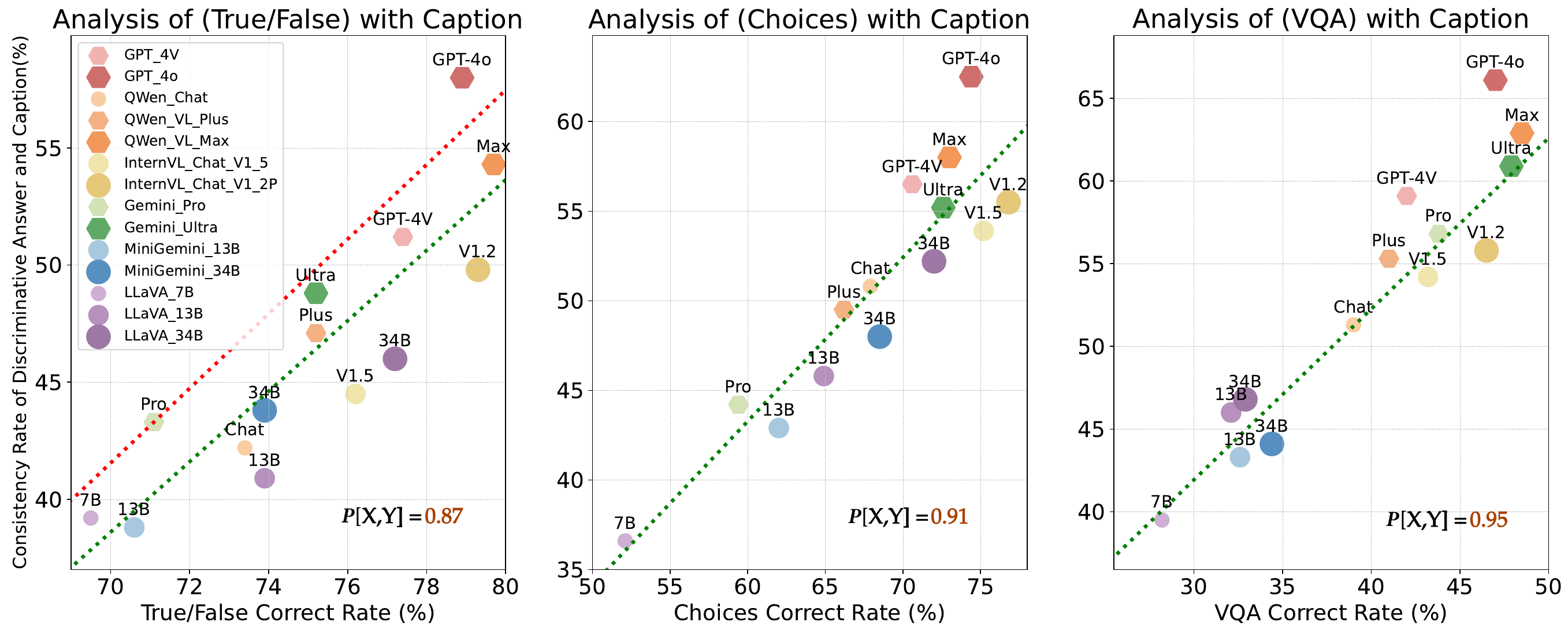}
        \vspace{-4mm}
        \caption{Visualization of relationship between correct rate of discriminative answer and its Consistency with caption on different \textbf{answer types}. From left to right, we visualize the following: True/False, multiple-choice and VQA questions.}
        \label{fig:all_3}
        \vspace{-2mm}
    \end{figure}
    
    \begin{figure}[t]
        \centering
        \includegraphics[width=1\textwidth]{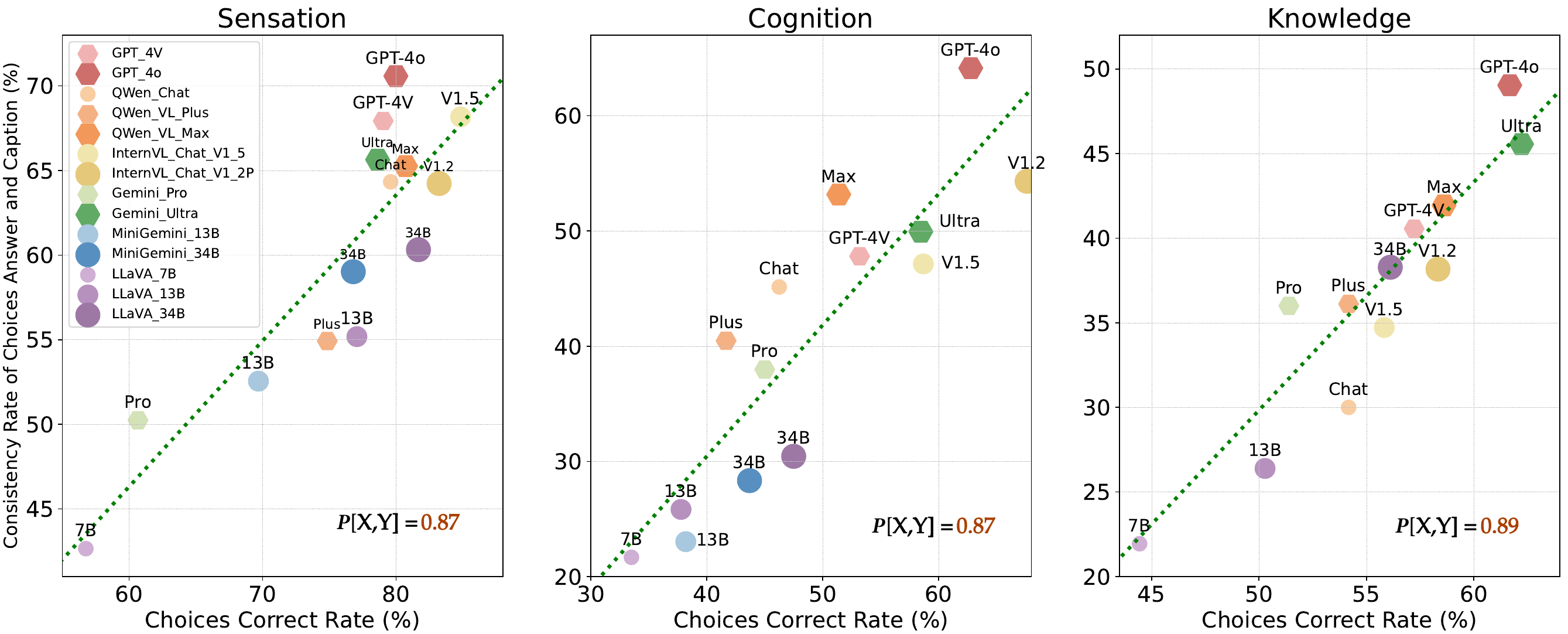}
        \vspace{-4mm}
        \caption{Visualization of relationship between correct rate of discriminative answer and its Consistency with caption on different \textbf{capability types}. From left to right, we visualize the following: sensation, cognition and knowledge.}
        \label{fig:all_3_2}
        \vspace{-3mm}
    \end{figure}

    \begin{theorem}[Inconsistency to Generative Answers]
        As the solution space of discriminative questions increases, the Consistency between their answers and generative answers increases.
    \end{theorem}
        As indicated in the last column of Table \ref{tab:caption_board}, “Ordered” means Con[T] < Con[C] < Con[V]. The answers of all closed-source models and most open-source models adhere to this pattern. Here is the theoretical explanation.
        Assume the distribution for the generative domain (Caption) is $\mathcal{S}$, and the sample space of $\mathcal{S}$ is $W$. For the discriminative domain, the sample space is limited to $W'$, which only contains some candidates from $W$.  Assume the model handles the discriminative domain by creating another distribution $\mathcal{S'}$ according to $\mathcal{S}$ and $W'$. 
        Then the total variation distance (TVD) \cite{kennedy1989total} between $\mathcal{S}$ and $\mathcal{S'}$ is
        \begin{equation} \label{eq:2}
            \frac{1}{2} \left\| \mathcal{S} -\mathcal{S'} \right\|_1 .
        \end{equation}
        This becomes larger when $| W \setminus W' |$ becomes larger. For instance, if the model creates $\mathcal{S'}$ by simply doing reject sampling\footnote{e.g, projecting or clustering $\mathcal{S}$ to $W'$}, then
        \begin{equation} \label{eq:3}
            \frac{1}{2} \left\| \mathcal{S} -\mathcal{S'} \right\|_1 = Pr[\mathcal{S} \in W \setminus W'].
        \end{equation}
        It is obvious that when $W'$ is more ”different” from $W$, the distance will be larger.

    \begin{theorem}[Connection between Discriminative and Generative Domain]
        The accuracy of the discriminative answer exhibits a strong positive correlation with its Consistency with the generative.
    \end{theorem}
        As shown in Figure \ref{fig:all_3} and \ref{fig:all_3_2}, we conduct visualizations for all tested LVLMs: The vertical axis represents the accuracy of their discriminative answers, while the horizontal axis represents the consistency of the answers with caption. Figure \ref{fig:all_3} displays the distribution across different question types, while Figure \ref{fig:all_3_2} illustrates the distribution across different core capabilities. The green lines represent a fitted linear equation. Additionally, we utilize the Pearson coefficient $P[X,Y]$ to quantitatively analyze the degree of linear correlation, and the 6 coefficients in the above figures are all more than $0.85$.
        

\subsection{Consistency Bias}

    \begin{theorem}[Consistency Bias]
        Closed-source models exhibit a pronounced bias advantage on Consistency, compared to open-source models.
    \end{theorem}

        When we fit a linear regression to all evaluated models and get the green line in Figure \ref{fig:all_3} (a):
        \begin{equation} \label{eq:4}
            \mathcal{L}_1: y = kx + b,
        \end{equation}
        where $x$ is the accuracy, and $y$ means Consistency between its answer and caption.
        We found that the majority of open-source models lie below this line, while closed-source models lie above it. In other words, at the same level of accuracy, the responses from closed-source models tend to exhibit better consistency with their captions. So we fit a linear regression to closed-source models and get the red line. The line they reside on has a higher bias $b_c$ (e.g., $b_c - b = 3.24$ in Figure \ref{fig:all_3} (a)), which aligns with our experience where closed-source models provide more comprehensive and reliable answers.

\section{Trigger-based Diagnostic Refinement}
\label{improve}

\begin{table}[t]
    \centering
	\renewcommand\arraystretch{1.7}
	\setlength\tabcolsep{0.55mm}
	\caption{\textbf{Results on LLaVA-34B and MiniGemini-34B via Trigger-based Diagnostic Refinement.}}
	\vspace{-3mm}
	\label{tab:improve}
	\scriptsize
    \begin{center}
	\begin{tabular}{l|c|p{1.6cm}p{1.6cm}p{1.6cm}}
	    \shline
        \footnotesize{Method} & \footnotesize{ConScore[C]} & \footnotesize{Con[T]} & \footnotesize{Con[C]} & \footnotesize{Con[V]} \\
	    \shline
            LLaVA-NeXT-34B \cite{liu2024llavanext} & $48.3 $ & $46.00$ & $52.20$ & $\mathbf{46.80}$\\
            \rowcolor{mygray}
            \quad \quad + TDR & $\mathbf{57.4}$ $(\fg{9.1 \uparrow})$  & $\mathbf{69.10}$ & $\mathbf{57.40}$ & $45.70$\\
	    \shline
            MiniGemini-34B \cite{li2024mini} & $49.6$ & $56.80$ & $48.00$ & $44.10$ \\
            \rowcolor{mygray}
            \quad \quad + TDR & $\mathbf{60.2}$ $(\fg{9.6 \uparrow})$& $\mathbf{76.10}$ & $\mathbf{53.80}$ & $\mathbf{50.80}$ \\
	    \shline
	\end{tabular}
         \end{center}
    \hspace{2mm}
    \vspace{-4mm}
\end{table}

\begin{wrapfigure}{r}{0.43\textwidth}
    \vspace{-5mm}
    \begin{center}
        \includegraphics[width=\linewidth]{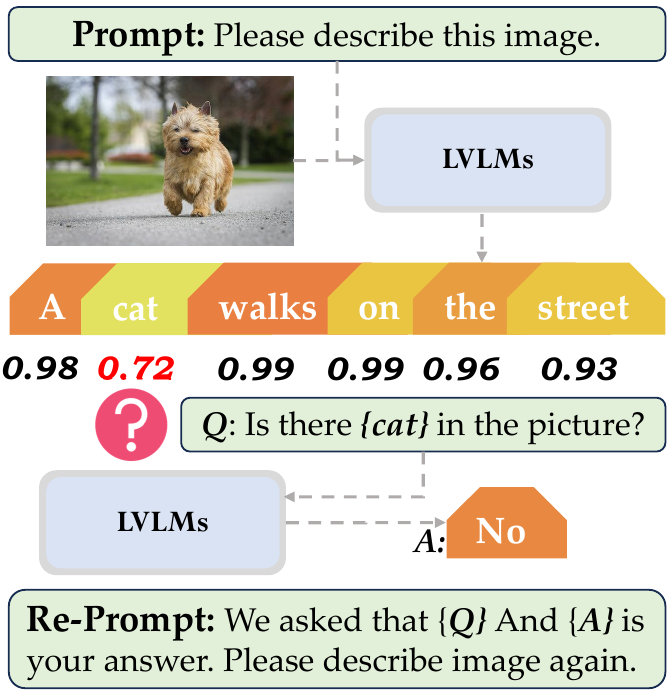}
    \end{center}
    \vspace{-4mm}
    \caption{\textbf{The Trigger-based Diagnostic Refinement pipeline.}}
    \label{fig:improve}
    \vspace{-5mm}
\end{wrapfigure}

In light of the previous findings, we summarize two key insights: (1) LVLMs exhibit higher accuracy when operating within a narrower discriminative solution space; (2) Incorrect answers are usually associated with significantly lower confidence and logits. Consequently, we propose a simple but efficient method dubbed Trigger-based Diagnostic Refinement (TDR) to ameliorate the generation skill of LVLMs without any additional training. The proposed pipeline is presented in Figure \ref{fig:improve}.

\textbf{Method} 
We start by making the LVLM generate a caption, with each word accompanied by its corresponding probability. Next, uninformative words are dropped based on their parts of speech, and we only keep nouns, adjectives and quantifiers. When the remaining words with probabilities below a threshold $\tau$ (we set $\tau$ = 0.85 here), trigger subsequent diagnostic processes. Since low probabilities of words indicate a lack of confidence, we formulate True/False discriminative questions to force the LVLM to self-verify (e.g., \verb+Is there {cat} in the picture?+). The self-diagnostic prompt and its response will be drafted into a new prompt, which is fed back into the LVLM to generate a higher-quality caption.

\textbf{Results} 
We carried out experiments on the LLaVA-NeXT-34B and MiniGemini-34B and evaluated them on the metric[C] of ConBench. The experimental results are detailed in Table \ref{tab:improve}.
Notably, the LLaVA-NeXT-34B sees an improvement of $9.1$ points, while the MiniGemini experiences an overall enhancement of $9.6$ points. Although our approach primarily employs True/False questions for self-verify, there is still a noticeable improvement in ConScore[C].
Hence, our method effectively boosts the quality of captions by triggering the model to self-check.

In theory, we can further construct multiple discriminative questions for the caption, enabling the model to verify multiple elements within the caption. Additionally, the process can be iterated multiple rounds, leading to ongoing enhancements in the quality of the generated output. Our method is a simplified implementation of the above approaches.

\section{Conclusion}

In this study, we investigate the Consistency issues in large vision-language models (LVLMs). Consistency reflects the overall ability of LVLMs, as it not only requires LVLMs to provide correct answers but also demands sufficient confidence in their knowledge point, regardless of the type of question encountered. We first introduce the ConBench, a benchmark that fills the gap in assessing Consistency. It includes 1K images with 4K prompts and two evaluation metrics: ConScore[D] and ConScore[C]. Then, our findings shed light on the nature of Consistency in LVLMs according to the ConBench. We observe that as the solution space of a prompt increases, the accuracy of the answers tends to decrease. Besides, we establish a relationship between the discriminative and generative realms, highlighting the importance of Consistency between the discriminative answer and caption. Furthermore, we discover that closed-source models exhibit a bias advantage over open-source models in terms of consistency. Finally, we propose a solution by forcing LVLMs to self-think, where a discriminative prompt is constructed via uncertain words in the caption. Our method makes the quality of LVLMs’ captions an impressive achievement. We believe that our research contributes to the evaluation of LVLMs and encourages future advancements for achieving Consistency in LVLMs. 

\newpage
{
\small
\bibliography{main}
\bibliographystyle{abbrv}
}

\newpage
\newcommand{\fgr}[1]{\textbf{\textcolor{ForestGreen}{#1}}}
\newcommand{\fred}[1]{\textbf{\textcolor{Forestred}{#1}}}

\appendix
\section*{Appendix}

\section{Showcases of Trigger-based Diagnostic Refinement}

\textbf{MGM-34B}

\begin{wrapfigure}{l}{0.3\textwidth}
    \vspace{-5mm}
    \begin{center}
        \includegraphics[width=\linewidth]{fig/000000236721.jpg}
    \end{center}
    \vspace{-4mm}
    \vspace{-5mm}
\end{wrapfigure}

\textbf{Image.} Count-000000236721.jpg

\textbf{Original Caption.} \textit{The image contains a variety of objects, which can be counted as follows: one tray, one cake, one bowl of sauce, one carton of milk, \fred{one banana}, one glass of orange juice, \fred{one cup of coffee}, and one sugar packet. The objects are arranged neatly on the tray, which is placed on a table.}

\textbf{Refined Caption} \textit{The image contains a total of \fgr{two bananas}, which are positioned on the right side of the tray. The bananas are yellow, indicating that they are ripe, and they are the only fruits visible in the image. They are placed next to a carton of milk, ice cream, a cake, \fgr{two cups of coffee}, a glass of orange juice, which are a common accompaniment in a breakfast setting.}

\textbf{LLava-Next-34B}
\begin{wrapfigure}{l}{0.3\textwidth}
    \vspace{-1mm}
    \begin{center}
        \includegraphics[width=\linewidth]{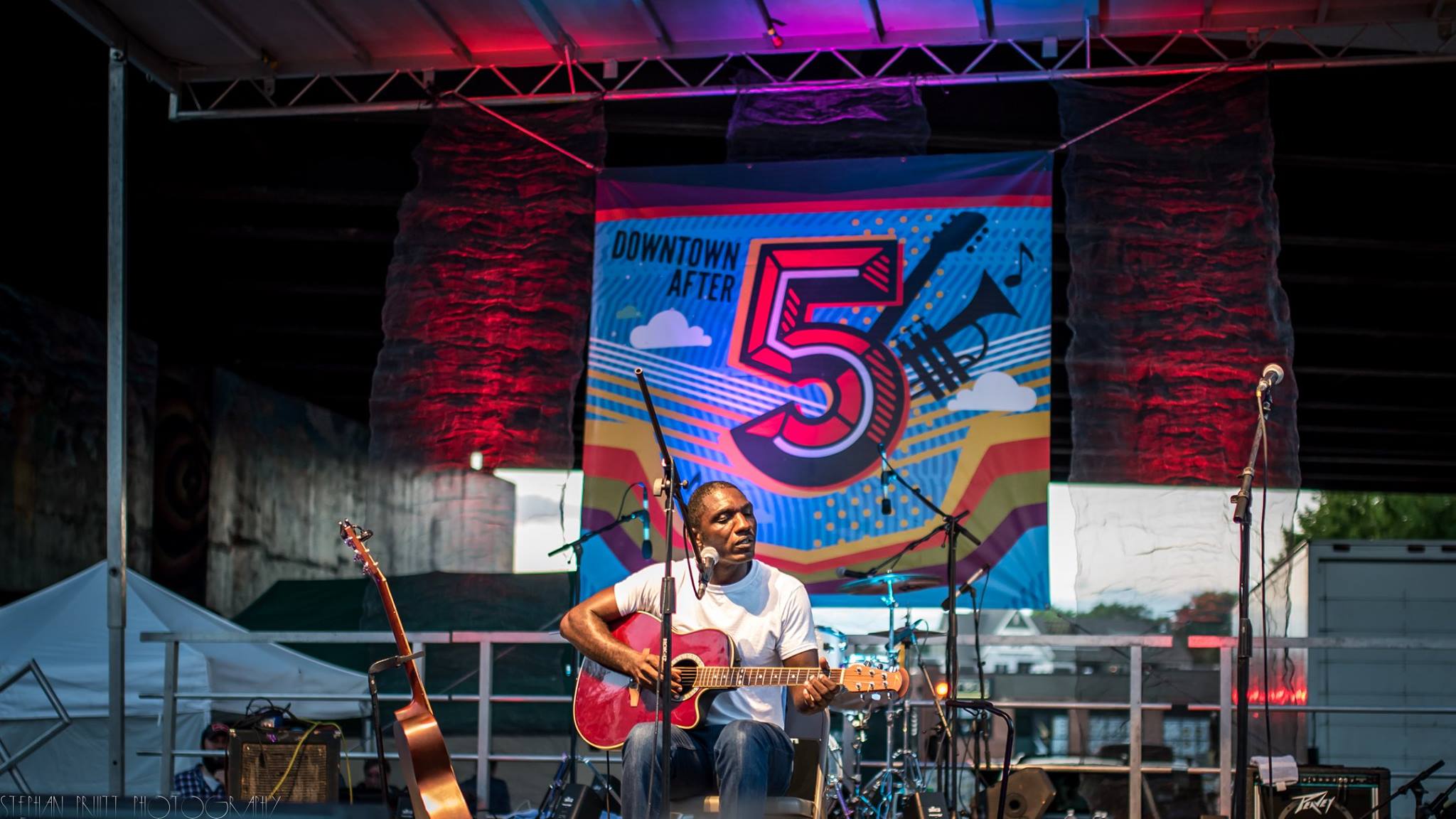}
    \end{center}
    \vspace{-4mm}
    \vspace{-5mm}
\end{wrapfigure}

\textbf{Image.} Attribute Reason-1724416\_2902943546.jpg

\textbf{Original Caption.} \textit{The image captures a vibrant live music scene. At the center of the frame, a musician is seated on a stool, engrossed in playing a red electric guitar. He is dressed casually in a white t-shirt and blue jeans, adding to the relaxed atmosphere of the performance. The stage, adorned with a colorful banner, is set against a backdrop of a brick wall, lending an urban feel to the setting. The banner prominently features the text "DOWNTOWN 5", possibly indicating the name of the band or the event. The musician is \fred{surrounded by other band members, each engrossed in their own instruments}, contributing to the overall harmony of the performance. The stage is well-equipped with a microphone and a speaker, essential for amplifying the sound and reaching the audience. In the background, a crowd of people can be seen, their attention riveted on the performance.}

\textbf{Refined Caption} \textit{The image captures a vibrant scene of a live music performance. At the center of the stage, a musician is seated on a stool, engrossed in playing an electric guitar. He is dressed in a casual white t-shirt, and his focus is evident as he strums the strings. The stage is set against a backdrop of a colorful banner that proudly proclaims "DOWNTOWN 5". The banner's bold colors contrast with the musician's white t-shirt, drawing attention to both the performer and the event. The stage \fgr{is equipped with various musical instruments}, including a drum set and a keyboard, suggesting a full band performance. A microphone stands ready, capturing the musician's performance for the audience. The stage is situated in a space that appears to be partially open to the elements, as suggested by the presence of a roof structure in the background. This setup indicates that the concert is being held in a venue that can accommodate outdoor or semi-outdoor performances, adding a unique ambiance to the event. Overall, the image portrays a dynamic and engaging live music event, with the musician at the heart of the performance, surrounded by his instruments.}

\section{Computation for Probabilities and Logits}
\label{stat}

For each model (MGM-13B and LLaVA-13B), we randomly selected 50 consistent and 50 inconsistent samples, respectively. The JSON samples are in support materials. To measure the LVLMs' response confidence accurately, we manually simplified the answers and maintained the words that are the most critical to the judgment. The specific principles are as listed:

For true/false questions, we only keep the word 'yes' or 'no' and their probabilities.
\begin{center}
    \verb+e.g., [Yes], there is a cat. +
\end{center}

For multiple-choice questions, we only keep the choice labels (e.g., A, B, C, D) and their probabilities.
\begin{center}
    \verb+e.g., ([A]) Cats. +
\end{center}

For limited VQA questions, we manually picked out keywords that matched ground truth from the answers, and computed the average probabilities of them as the final probability.
\begin{center}
    \verb+e.g., A [Cat] walks on the street.+
\end{center}

\section{Limitations}
\label{Limitations}
The introduced ConBench offers a new perspective on evaluating model performance through the consistency between multiple types of questions, providing a more comprehensive measurement and understanding of existing LVLMs. However, due to the distinct response forms of captions, assessing the consistency between captions and discriminative answers is judged by GPT, posing a risk of inaccurate evaluations. Besides, by delving deeper into our benchmark analysis, we propose trigger-based diagnostic refinement to improve the consistency and accuracy of LVLMs. This, however, introduces additional computational costs and is limited by the inherent capabilities of the LVLMs. Further improvements can be achieved by designing and training LVLMs with a focus on consistency.


\section{Broader Impacts}
\label{Impacts}

Overall, this research has broader impacts on the evaluation, performance, fairness, and future development of LVLMs, fostering progress and advancements in the field of vision-language models.

\textbf{Advancing Evaluation: }The introduction of ConBench, a benchmark for assessing Consistency in LVLMs, fills a crucial gap in the evaluation of these models. This benchmark provides a standardized framework for measuring the performance and reliability of LVLMs across different prompts.

\textbf{Novel Insights:} we are the first to reveal the tapestry and get the following findings:
(1) In the discriminate realm, the larger the solution space of the prompt, the lower the accuracy of the answers. 
(2) Establish the relationship between the discriminative and generative realms.
(3) Compared to open-source models, closed-source models exhibit a bias advantage in terms of Consistency.

\textbf{Inspiring Future Research:} By contributing to the evaluation and understanding of Consistency in LVLMs, this research paves the way for future advancements in the field. It encourages researchers to explore new techniques, methodologies, and approaches to achieve higher levels of Consistency in LVLMs, ultimately pushing the boundaries of language and vision understanding.

\section{Detailed Cases in ConBench}
We have uploaded the ConBench dataset, including images and their prompts, to the Hugging Face platform. The dataset can be accessed at the following URL: \url{https://huggingface.co/datasets/ConBench/ConBench}. Here, we enumerate several representative cases from ConBench. Arrange in order from easy to difficult, respectively, based on sensation, cognition, and knowledge.

\afterpage{
  \clearpage
  \begin{figure}[p]
    \centering
    \includegraphics[width=1\textwidth]{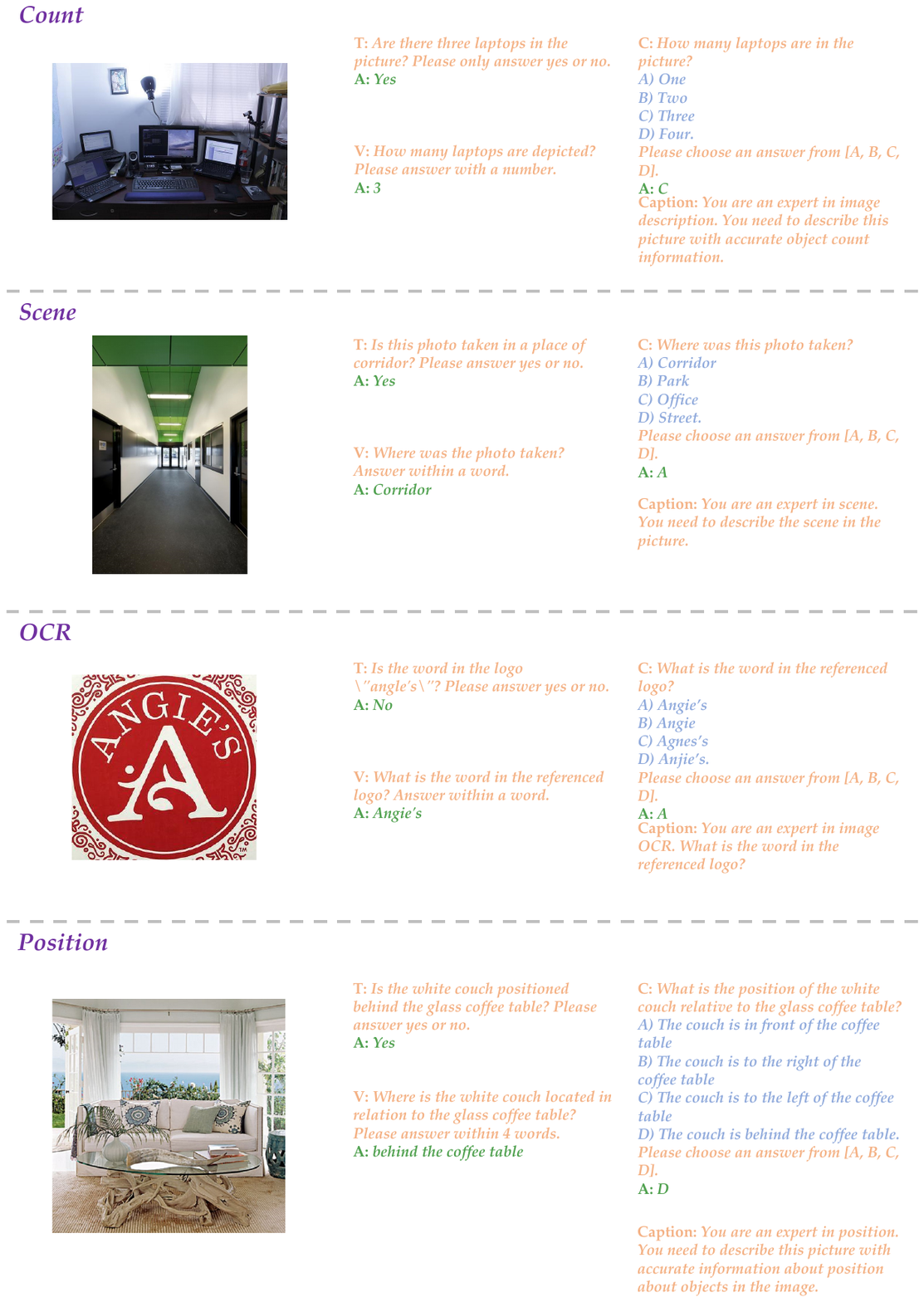}
    \vspace{-4mm}
    \label{fig:V1}
    \vspace{-2mm}
  \end{figure}
  \clearpage
}

\afterpage{
  \clearpage
  \begin{figure}[p]
    \centering
    \includegraphics[width=1\textwidth]{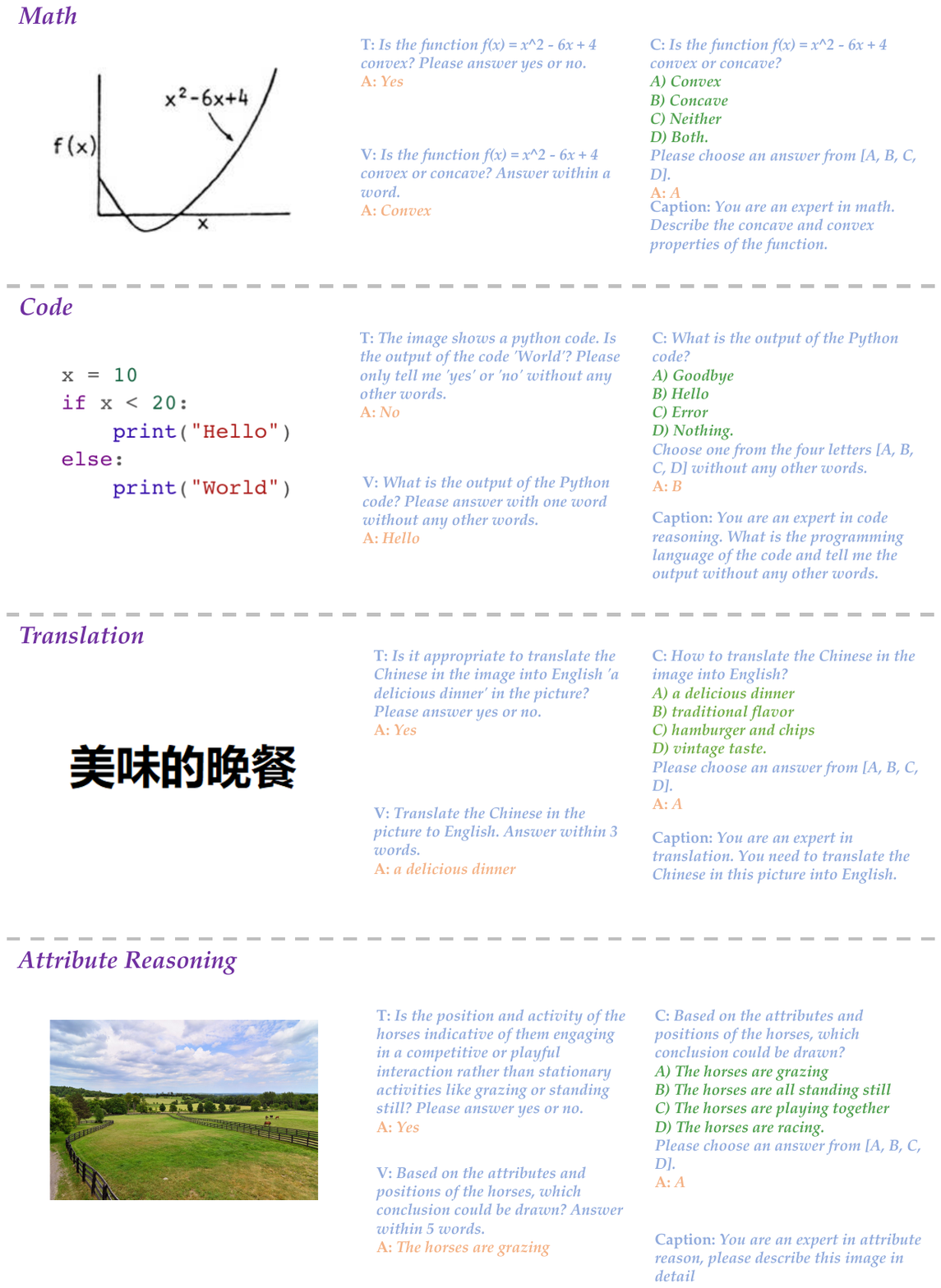}
    \vspace{-4mm}
    \label{fig:V2}
    \vspace{-2mm}
  \end{figure}
  \clearpage
}

\afterpage{
  \clearpage
  \begin{figure}[p]
    \centering
    \includegraphics[width=1\textwidth]{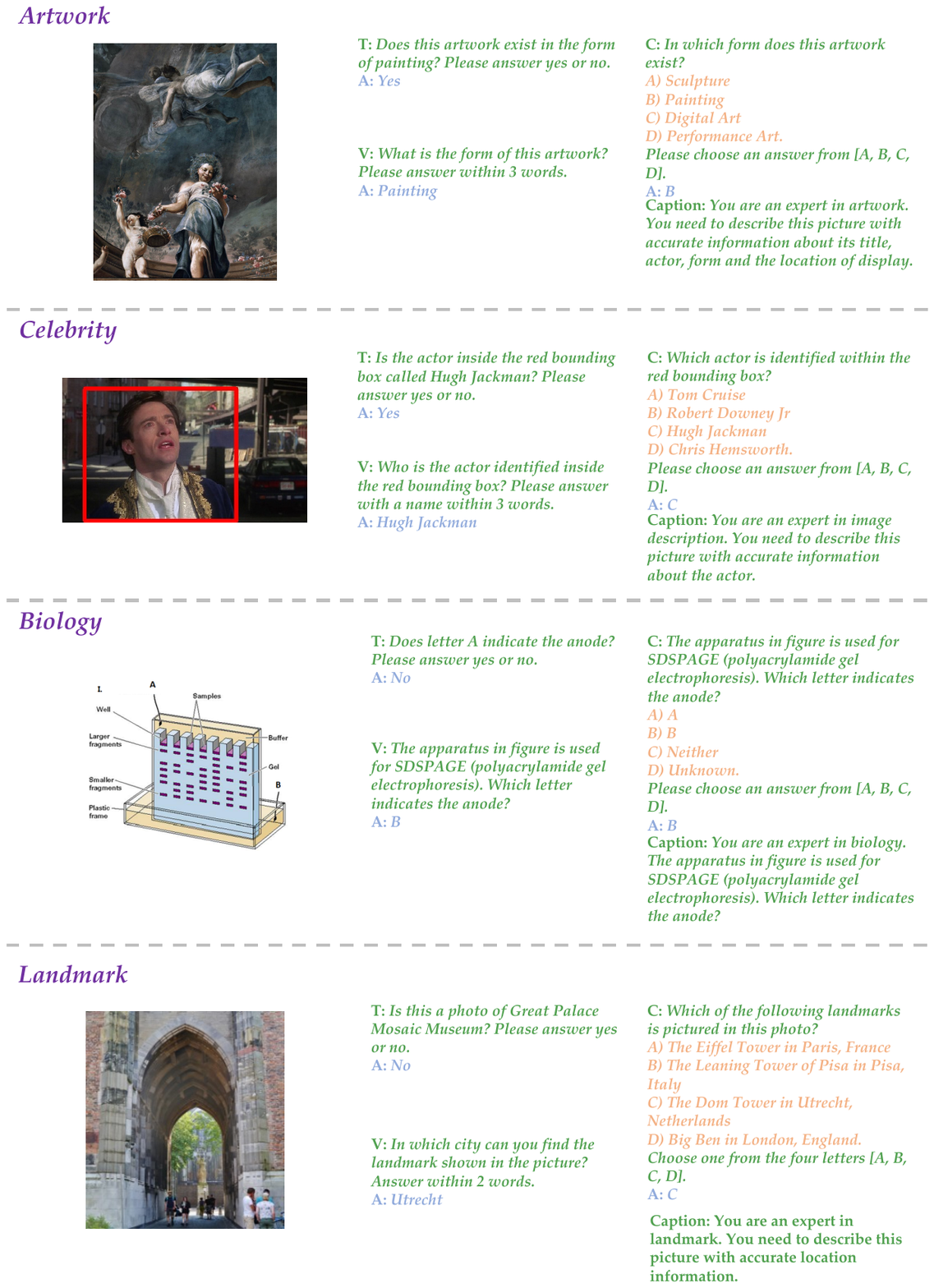}
    \vspace{-4mm}
    \label{fig:V3}
    \vspace{-2mm}
  \end{figure}
  \clearpage
}


\end{document}